\documentclass{bmvc2k}

\title{BrainSegNet : A Segmentation Network for Human Brain Fiber Tractography Data into Anatomically Meaningful Clusters}

\addauthor{Tushar  Gupta}{tushar.behror@gmail.com}{1}
\addauthor{Shreyas Malakarjun Patil}{patil.3@iitj.ac.in}{2}
\addauthor{Mukkaram Tailor}{mukarram.tailor02@gmail.com}{1}
\addauthor{Daksh Thapar}{s16007@students.iitmandi.ac.in}{1}
\addauthor{Aditya Nigam}{aditya@iitmandi.ac.in}{1}
\addinstitution{
 School of Computing and Electrical Engineering, \\
 Indian Institute of Technology, Mandi\\
 Mandi, HP, India
}
\addinstitution{
 Department of Electrical Engineering, \\
 Indian Institute of Technology, Jodhpur\\
 Jodhpur, Rajasthan, India
}

\runninghead{BrainSegNet}{Brain Tractography Segmentation Network}

\begin{document}

\maketitle

\begin{abstract}
The segregation of brain fiber tractography data into distinct and anatomically meaningful clusters can help to  comprehend the complex brain structure and early investigation and management of various neural disorders. We propose a novel stacked bidirectional long short-term memory(LSTM) based segmentation network, (BrainSegNet) for human brain fiber tractography data classification. We perform a two-level hierarchical classification a) White vs Grey matter (Macro) and b) White matter clusters (Micro). BrainSegNet is trained over three brain tractography data having over 250,000 fibers each. Our experimental evaluation shows that our model achieves state-of-the-art results. We have performed inter as well as intra class testing over three patient's brain tractography data and achieved a high classification accuracy for both macro and micro levels both under intra as well as inter brain testing scenario.
\end{abstract}

\section{Introduction}
\begin{figure}
\centering
\subfigure[Brain Slice]{\includegraphics[width=0.14\linewidth, height=0.2\linewidth]{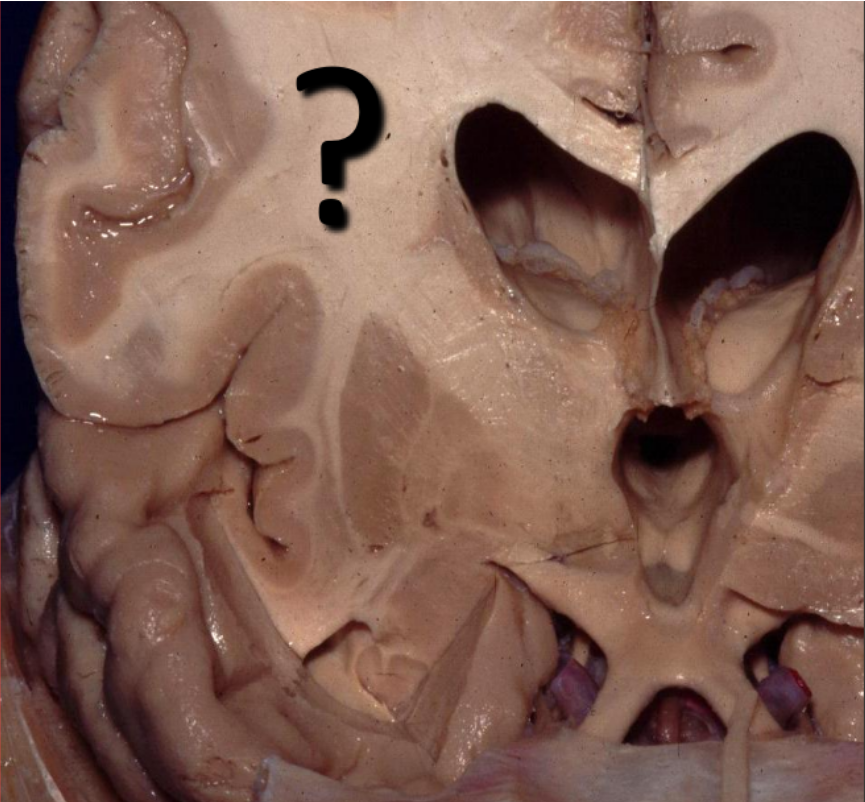} \label{fig:1}}
\subfigure[Diffusion]{\includegraphics[width=0.16\linewidth, height=0.2\linewidth]{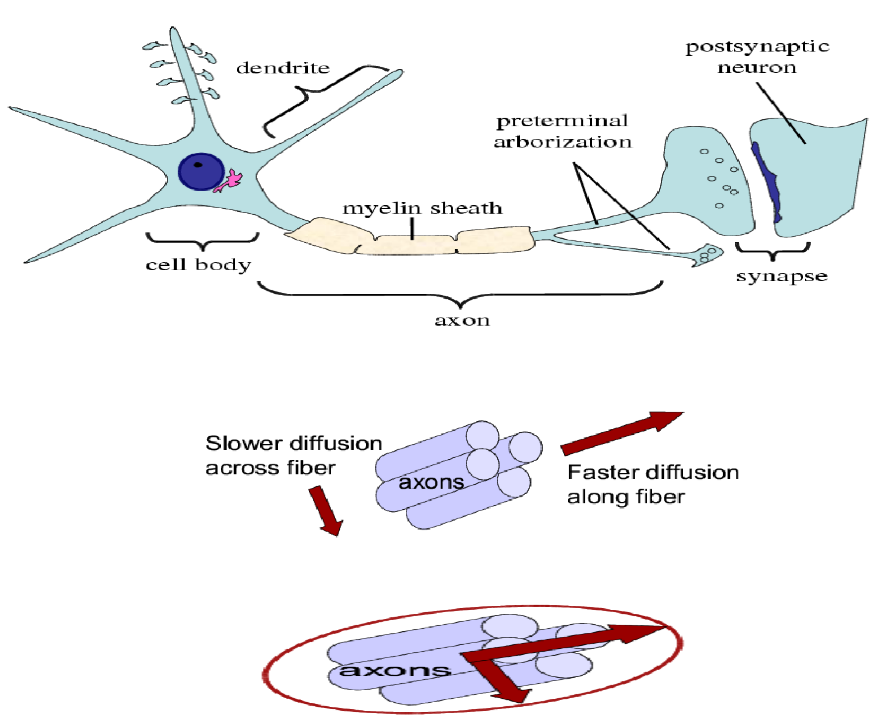}\label{fig:2}}
\subfigure[Directions]{\includegraphics[width=0.16\linewidth, height=0.2\linewidth]{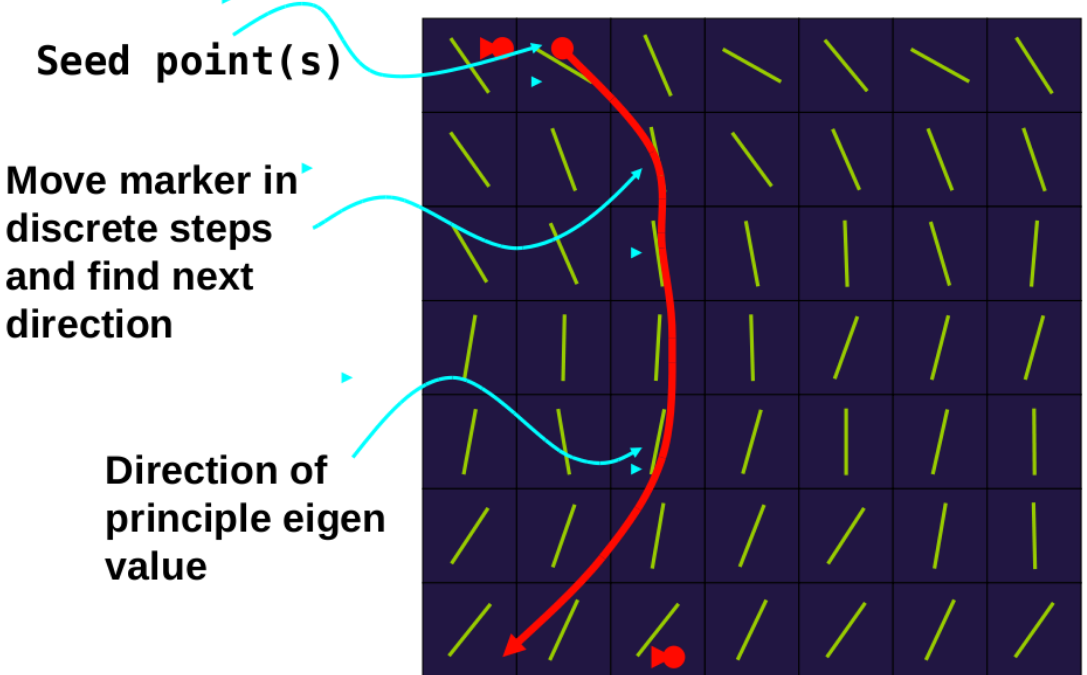}\label{fig:3}}
\subfigure[Fiber Tracking]{\includegraphics[width=0.18\linewidth, height=0.2\linewidth]{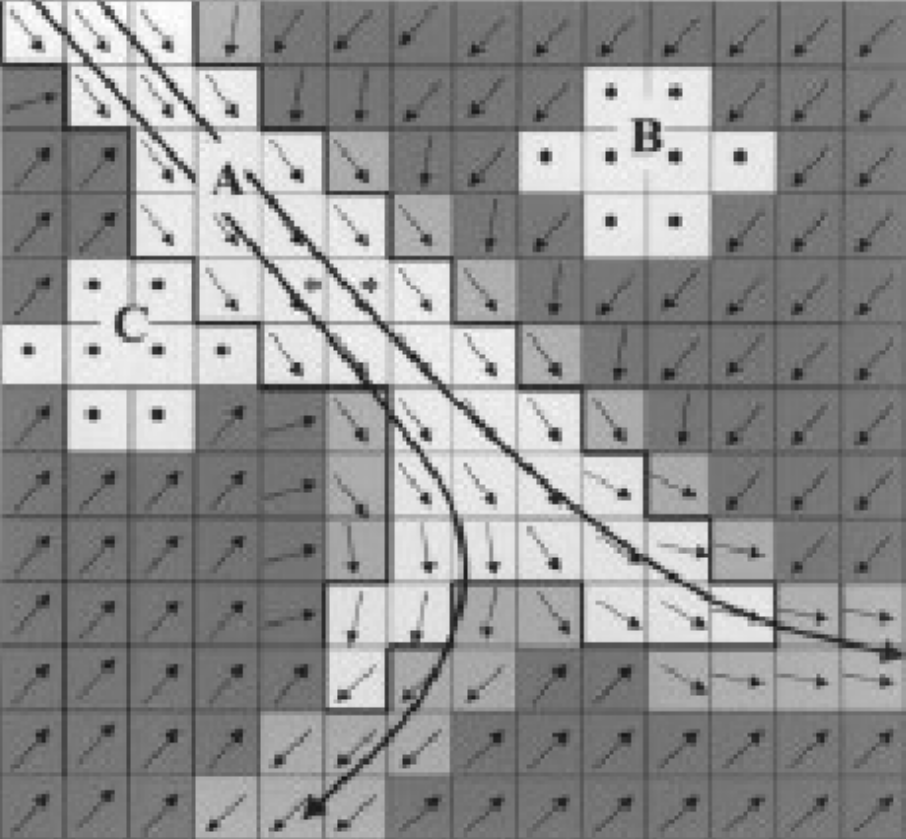}\label{fig:4}}
\subfigure[Tractography]{\includegraphics[width=0.16\linewidth, height=0.2\linewidth]{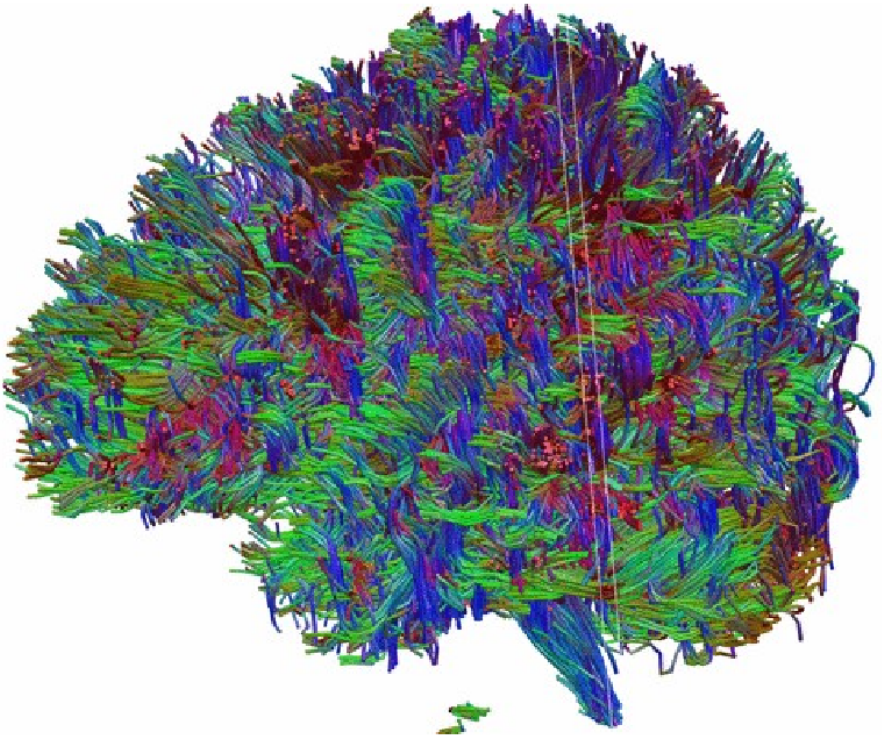}\label{fig:5}}
\subfigure[Segmentation]{\includegraphics[width=0.16\linewidth, height=0.2\linewidth]{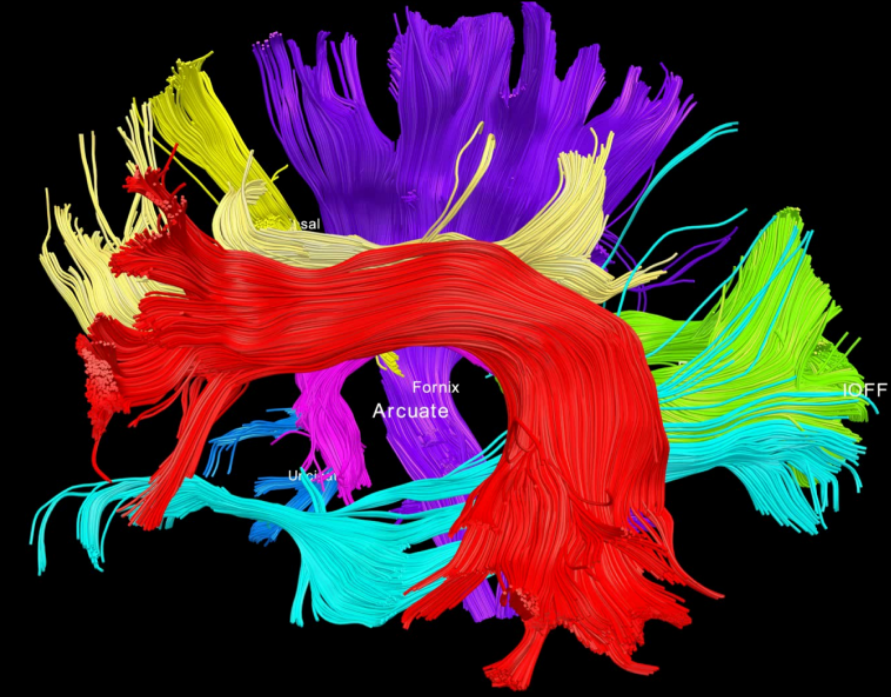}\label{fig:6}}
\caption{Images are taken from ~\cite{fernandez2012high},~\cite{guise2016hollow}}
\label{fig:init}
\end{figure}

The brain is the central processing unit of our body which maintains order and controls the actions of all the organs. Naturally, brain also communicates between its own subdivisions (active regions) for which it uses neuronal connections (termed as neuronal fibers), which consist of dendrites and axons. The dendrites serve as the receivers in a neuron and axons in any neuron are responsible for transmission of signals to other neurons. Typically a brain consist of several billions of such neurons constantly receiving and transmitting signals between themselves, forming a very complex network that commands our day to day activities. The brain fibers are broadly divided into two classes, namely the grey and white fibers as shown in Fig.~\ref{fig:1}. The white fibers mainly consist of axons that connect various parts of grey matter while the gray matter contains cell bodies, dendrites and axons as shown in Fig.~\ref{fig:2}. The white matter can be further classified into eight clusters - Arcute, Cingulum, Corticospinal, Forceps Major, Fornix, Inferior Occipitofrontal Fasciculus, Superior Longitudinal Fasciculus and Uncinate. These are basically bundles or clusters of fibers called neural tracts that form the pathways between different hemispheres and brain regions as shown in Fig.~\ref{fig:6}. 

The brain fiber tractography data can be extracted from 3T Magnetic resonance imaging(MRI) data using diffusion tensor imaging which is a non-invasive MRI technique. In this technique diffusion has been considered as the molecular fluid spread and its extent depends on the diffusive property of the medium as shown in Fig.~\ref{fig:3}. In brain white matter, a tissue consists of bundles of myelination axons. Crucially, it is more hindered across than along such bundles. Hence, by measuring diffusion along many directions and observing that it is faster in one direction than in others, one can deduce the direction of fiber bundles~\cite{guise2016hollow}. The process known as tractography, helps in visualizing the fiber structure in three dimensions using Diffusion Tensor Imaging(DTI) as shown in Fig.~\ref{fig:5}. One such example has been shown in Fig.~\ref{fig:mod}, where on the top-right each fiber is coloured randomly, for better visualization. The eight major fiber clusters are shown in bottom right, while the grey matter fibers are shown in bottom-left. Any fiber can be represented as a set of points in 3D space (typically 20 to 100 points per fiber). Tractography produces thousands of fiber trajectories per subject (around 250K). Finally, obtained data can be seen as a 3D point cloud that does not convey anything useful. In order to extract any useful information the fibers must have to be organized into anatomically meaningful structure. DTI may improve preservation of eloquent regions during surgery by providing access to direct connectivity information between functional regions of the brain, and it has progressively been incorporated into strategic planning for resection of complex brain lesions~\cite{fernandez2012high}.

\subsection{Problem Statement} 

\textbf{Designing brain tractography segmentation network (BrainSegNet) : } Given tractography data of human brain, we have designed a moderately deep recurrent neural network that can automatically segment brain fibers into tracts having ``similar'' fibers which are anatomically meaningful. Medically it is also believed that at coarser level there are two classes grey and white matter (Macro Level) and further in white matter their are eight cluster/tracts (Micro Level) \emph{viz.} \emph{Arcute, Cingulum, Corticospinal, Forceps Major, Fornix, Inferior Occipitofrontal Fasciculus, Superior Longitudinal Fasciculus} and \emph{Uncinate}.

\subsection{Motivation} 

Classification of these brain fibers provides a better understanding of the brain structure and also can be very helpful for planning brain tumor surgeries and other medical tasks involving the alterations/cutting of brain fibers. Also a much more detailed and systematic analysis can be planned over the brains suffering from many complex disease like Parkinson and Alzheimer, for which still not much is unraveled. The manual segmentation of various brain fiber is a very tedious task as the data available is enormous and a great amount of expertise is required. Therefore, an automated classification of these fibers into their respective anatomically meaningful clusters, is medically very important and a challenging problem.

\begin{figure}
\centering
\includegraphics[width=0.9\linewidth, height=0.5\linewidth]{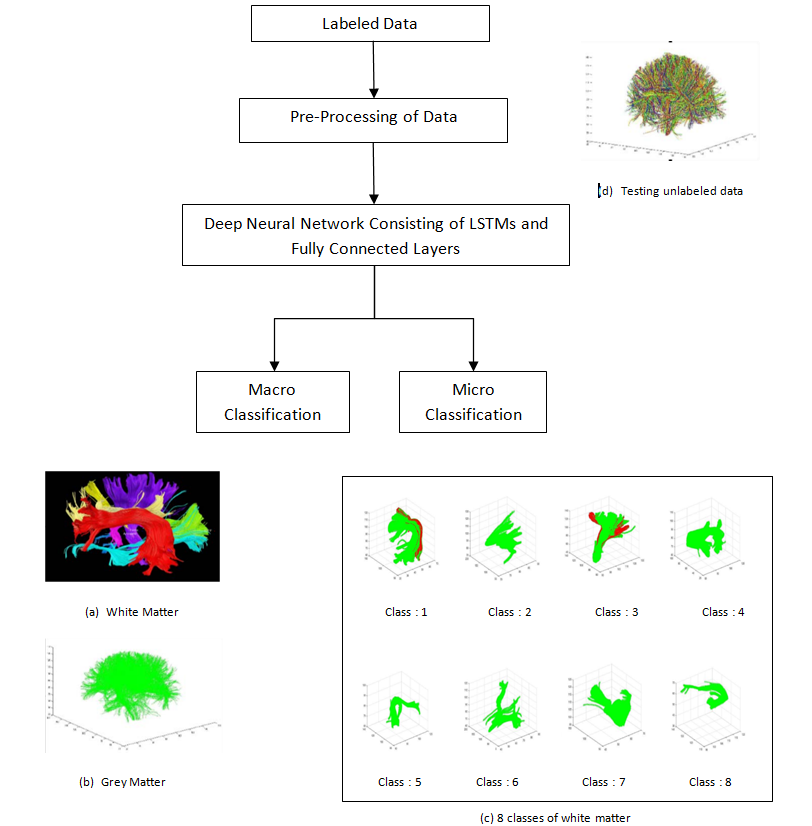}
\caption{Overview of the proposed model }
\label{fig:mod}
\end{figure}

The BrainSegNet model proposed in this paper, uses labeled DTI data for learning the features required to classify the individual brain fibers. It has been observed that the inter tract fibers follow different routes or paths from one part of the brain to another and paths are often non-linear. While fibers from the same class usually connect ``similar'' parts of brains and follow similar routes. One important feature we have used in this work is that we first extract a fixed fraction of points having the highest curvature and then use a network consisting of multiple LSTM layers and one bilateral LSTM layer at the beginning so as to funnel out the information from the gradients in both the positive as well as the negative network propagation direction. A two level hierarchical classification has been performed. One at the ``Micro''  level,  in which we  classify  eight primary classes  of  the  white  matter (WM) and  another at ``Macro'' level, in which binary classification  between two classes \emph{viz.} white matter(WM) and grey matter(GM) tracts. Any test fiber is classified into one of the classes by comparing a proximity based on the learned high-curvature-point based BrainSegNet model.

\subsection{Contribution} 
An entirely new model(BrainSegNet) has been formulated using bilateral LSTMs and LSTMs, which is in accordance with the theory of curvature points specifying the routes being the distinctive factors. The trained model when tested, surpassed the previous state of the art results.

\subsection{Previous Work} 
Although the problem of brain fiber classification is quite a recent development, still several unsupervised techniques have already been suggested ~\cite{catani2002virtual}, ~\cite{maddah2005automated} along with few supervised approaches. Majorly the unsupervised approaches manually select the Region of Interest (ROI) and then group the fibers through these ROIs. In~\cite{o2007automatic}, authors used spectral clustering to generate a white matter atlas automatically.  The similarity is calculated between fibers using Hausdorff distance, and the clustering is employed in embedded space, that is formed using the Eigen-vectors of the distance matrix.  In~\cite{wang2011tractography}, hierarchical Dirichlet process has been used to determine the number of clusters. Another supervised approach presented in~\cite{Patel:2016:ABT:3009977.3010013}, selects few major points  having maximum curvature and fed them to a clustering algorithm which takes into account the position and the curvature values while classifying.  


\section{Proposed Model : BrainSegNet}

In this section we discuss the various components of our model, the overall structure, data format, data pre-processing ($25\%$ of data pruning) and the detailed architecture of the proposed network consisting of LSTMs and bilateral LSTMs layers. The overview of our proposed model has been shown in Fig.~\ref{fig:mod}. 

\subsection{Tractography Data and its Format}
The full data of three patients brain and their respective labels has been taken from University of Pittsburg as a part of contest. The ground truth had been manually annotated by expert neurologists and surgeons using their own interactive tools. The data has been provided as $.trk$ or track file format. A track file is a single binary file, with the first $1000$ bytes are header information and the rest constitute the required fiber information. Each patient data consists of about $250,000$ fibers. The average number of sample points over a fiber across different classes varies between $36$ to $120$ highlighting that our approach is insensitive to the fiber length. Each of the fiber is labelled an integer in the range $0$ to $8$ with respect to $9$ classes, where there $1$ is used for grey matter and remaining $8$ different types of white matter classes/tracts. The number of fibers per class in the training set considered  for all the 3 patients has been depicted in Table~\ref{tab:1}.

\begin{table}[!h]
\begin{center}
\begin{tabular}{ |c|c|c|c|c|c|c|c|c|c|}
\hline
\textbf{Patient ID}&\textbf{Gray Matter}&\multicolumn{8}{|c|}{\textbf{White Matter}}\\
\hline
&C-0&C-1&C-2&C-3&C-4&C-5&C-6&C-7&C-8\\
\hline
B1&74,486&900&247&3,428&570&358&120&274&111\\
\hline
B2&75,384&462&868&1,937&374&246&497&107&125\\
\hline
B3&74,834&525&782&2,225&788&116&618&46&66\\
\hline
\end{tabular}
\end{center}
\caption{Number of fibres of different classes in $3$ brains considered in this work}
\label{tab:1}
\end{table}

\subsection{Curvature based Fiber Data Pruning using Partial Derivatives}

Fiber pruning has been done in two steps : a) Extracting meaningful data points - high curvature points, and b) Conversion into fixed input using masking. The labelled data of 3 brains is available to us in the form of .trk (Track) file format [Ref.], Track file is one single binary file, with the first 1000 bytes as the header and the rest as the body. 

\begin{enumerate} 
\item \textbf{Fiber Pruning : } Extraction of meaningful data points has been done under the assumption that similar fibers follow similar paths from one part brain region to another and hence would have almost ``similar'' curvature points. Only those sample points involved in high curvature are selected by taking projections of these vectors on each of the planes ($XY, YZ$ and $ZX$). Then accumulate their gradients in both directions with respect to the sample points just before and after them and as well as with respect to points, four steps ahead and behind from the current point. This scheme is chosen in order to handle the multi-scale curvature. Finally gradients are sorted and bottom $25\%$ points are pruned.

\item \textbf{Fixed Length Conversion : } The brain fibers are represented by a variable length sequence of points in a three dimensional space (3D vectors). Since our network requires fixed length input, the length of sequences was restricted to $100$ points. Longer sequences has been truncated while shorter ones are padded with zeros. This strategy lead to poor results as the fiber structure severely got affected. Later, masking layer was used instead of padding to preserve the structure of fibers. Masking skips a time-step where all features are equal to the mask value. 
\end{enumerate}

\textbf{Justification : } To evaluate pre-processing, we have tested our model over both pruned and original fiber and have observed that with longer sizes the training time increases while accuracy does not improve.

\begin{figure}
\centering
\includegraphics[width=0.8\linewidth, height=0.6\linewidth]{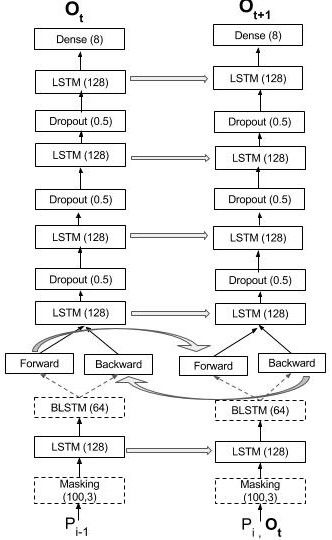}
\caption{ BrainSegNet Model Architecture}
\label{fig:arc}
\end{figure}

\subsection{Model Architecture}
Deep Neural Networks (DNNs) are powerful models that have achieved excellent performance on very complex learning tasks. We have used a deep stacked Long Short Term Memory (LSTM) network as shown in Fig.~\ref{fig:arc}. Histotically RNNs~\cite{sutskever2013training}  are built to utilize past information to predict the future using hidden layers. But RNN's failed to train well due to vanishing gradient problem and later LSTM Networks has been proposed.

\subsubsection{RNN (Recurrent Neural Networks)} 
RNNs \cite{sutskever2013training}  are built in order to utilize past information to predict the future by using a loop into the hidden layers that pass information to their respective successors. But RNN's with long dependencies failed to train well due to vanishing gradient problem. In order to learn long and short both types of dependencies LSTM Networks has been proposed.

\subsubsection{Long Short Term Memory Networks (LSTM)}
The LSTMs \cite{Hochreiter:1997:LSM:1246443.1246450} are special RNNs capable of learning long term dependencies. At each iteration information passed to their respective successors, that learns which information appeared in the past is most relevant discard remaining. The core concept behind LSTMs is that they can manipulate the inputs to the cells from their predecessors by using  three gates as discussed below:

\begin{enumerate}
\item \textbf{Forget Gate : } Acts as a filter for the information flow from previous cells.

\begin{equation}
f_t = \sigma (W_f.[h_{t-1},x_t]+b_f)
\end{equation}

Here, $f_t$ learns what to forget, $h_{t-1}$ is previous cell state, $x_t$ is current input, $b_f$ is bias and $W_f$ is weight matrix learned during training.  

\item \textbf{Input Gate :} It selects what to store in the current cell.

\begin{equation}
i_t = \sigma (W_i.[h_{t-1},x_t]+b_i)\hspace{0.2cm}
\end{equation}
\begin{equation}
C_{i_t} =\tanh (W_C.[h_{t-1},x_t]+b_C)\hspace{0.2cm}
\end{equation}
\begin{equation}
C_t = f_t*C_{t-1} + i_t*C_{i_t} 
\end{equation}

Here, $W_i$, $W_C$ are weight matrices,  $Ci_t$ is vector of new candidate values, $b_i$ and $b_C$ are biases and $C_t$ is new cell state.


\item \textbf{Output Gate : } Computes the output of the current state.
\begin{equation}
o_t = \sigma (W_o.[h_{t-1},x_t] + b_o);\hspace{0.2cm}
h_t = o_t * \tanh(C_t)
\end{equation}
Here, $W_o$ is weight matrices, $o_t$ learns significant cell state part, $h_t$ is the output of current hidden layer and  $b_o$ is outptut bias.

\end{enumerate}

\subsubsection{Justification for the selected model} 
We have observed that RNNs need to be deep enough to capture subtle differences in structure of fibres from different classes. Initial few layers consist of one Bi-directional LSTM to capture structural understanding at each fiber point and its curvature characteristics. The next $3$ are LSTM layers with reducing memory units as shown in Fig.~\ref{fig:arc}. Finally, fully connected layers with a sigmoid activation function has been used to make predictions.

\subsubsection{Training} 
Deep stacked LSTMs often give better accuracy over shallower models. However, simply stacking more layers of LSTM works only to a certain number of layers, beyond which the network becomes too slow and difficult to train due to exploding and vanishing gradient problem~\cite{Hochreiter:1997:LSM:1246443.1246450}. We have observed that stacked LSTM layers work well up to $4$ layers, barely with $6$ layers, and very poorly beyond $8$ layers. The training set was split for validating and training at a ratio of $1:5$, so as to validate the training after each epoch. 

\

\noindent
\textbf{Network Hyper-Parameters : } Hyper parameter used in the model are as follows : 
\begin{itemize}
 \item Number of epochs : 15
 \item Batch Size : 64
 \item Activation Function : Sigmoid
 \item Optimizer Function : Adam
 \item Fraction of Data used for training : 0.4
 \item Loss functions : Binary Cross-entropy and Categorical Cross-entropy for macro and micro respectively
 \item Validation set of size 0.2 of training set size has been used.
\end{itemize}

\section{Experimental Analysis}
In this section we talk about the methods applied for validation of our proposed model. The major challenge that we have faced is to get more and more ``labeled tractography'' data. We some how got hold of three brain data from a contest organized by Pittsburg. Hence, we have done the validation over a dataset of three patients brain data containing various fibers which are represented by a variable length of sequence and each of the fiber is labeled to one of the 9 classes. Each brain data contains $250,000$ fibers and their respective class. We have formulated this problem as two level hierarchical classification problem in which experiments are carried first at, a) Macro Level - In which the fibers undergo binary classification with respect to two classes \emph{viz.} grey and the white matter (data imbalance is a big challenge), secondly  at  b) Micro Level - In which the fibers are classified into one of the 8 sub-classes of white fibers (intra-class variation is a big challenge).

\subsection{Testing Methodology} The testing of our trained BrainSegNet model has been done using three protocols defined below :

\begin{enumerate}
\item \textbf{Intra Brain Testing : } BrainSegNet has been trained and tested over the same patient's data (one of the three patient).
\item \textbf{Inter Brain Testing : } BrainSegNet has been trained over one of the three patient's data while tested on a fraction of the data from other two patients. We have reported our results only when trained over patient 2 (\emph{i.e.} Brain 2) data and tested over remaining Brain 1 and Brain 3 data.
\item \textbf{Merged Brain Testing : } BrainSegNet has been trained over merged data of all the three patients, such that half of the data points from each brain has been considered for training after shuffling, while remaining patients data has been shuffled and used for testing.
\end{enumerate}


All of the above mentioned testing strategies are performed for both Micro classification and macro classification as described. We have used two performance parameters (a) Accuracy and (b) Recall defined in the Eqs. \eqref{eq1}, \eqref{eq2}. Since the number of grey fiber is much more than that of white fibers, as shown in Table~\ref{tab:1}, we have computed recall only for white fibers as it would skew the results for grey fibers.


\begin{equation}\label{eq1}
\begin{split}
Accuracy & = \frac{\# \ correctly \hspace{0.1cm}classified\hspace{0.1cm} fibers}{Total\hspace{0.1cm}\hspace{0.1cm}number\hspace{0.1cm} of\hspace{0.1cm} fibers} \\
\end{split}
\end{equation}

\begin{equation}\label{eq2}
\begin{split}
Recall & = \frac{\# \ white\hspace{0.1cm} fibers \hspace{0.1cm}predicted}{Total\hspace{0.1cm} Number\hspace{0.1cm} of\hspace{0.1cm} White\hspace{0.1cm} fibers} \\
\end{split}
\end{equation}


\begin{table}[!h]
\begin{center}
\begin{tabular}{|c|c|c|c|c|c|c|}
\hline

 \textbf{Patient} & \multicolumn{4}{|c|}{\textbf{Accuracy}($\%$)}  & \multicolumn{2}{|c|}{\textbf{Recall}($\%$)} \\ 
 \hline
  &\multicolumn{2}{|c|}{\textbf{Macro}}&\multicolumn{2}{|c|}{\textbf{Micro}} &\multicolumn{2}{|c|}{\textbf{Macro}}\\  
 \hline
 \multicolumn{7}{|c|}{\textbf{Intra: Trained and tested over same brain}} \\
 \hline
 &ANN~\cite{Patel:2016:ABT:3009977.3010013}& BrainSegNet & ANN~\cite{Patel:2016:ABT:3009977.3010013}& BrainSegNet &  ANN~\cite{Patel:2016:ABT:3009977.3010013} & BrainSegNet \\
\hline
 \textbf{B1} &98.02 &98.88 &97.34& 98.12 &82.1& 94.98 \\
 \hline
 \textbf{B2} &96.14& 97.84 &93.49& 99.96 &65.2& 80.68 \\
 \hline
 \textbf{B3} &96.69& 96.42 &95.96& 97.45 &57.6& 73.45 \\
 \hline
 \multicolumn{7}{|c|}{\textbf{Inter: Trained on B2 and tested on B1 and B3}} \\
\hline
 \textbf{B1} & --- &93.30 & --- &99.55 &---&49.00 \\
 \hline
 \textbf{B3} & --- &94.47 &---&96.18 & --- &49.60 \\
 \hline
 \multicolumn{7}{|c|}{\textbf{Merged: Trained and tested over merged brain data}} \\
\hline
 &94.87&96.65 &93.95& 95.51 &80.6& 83.46 \\
 \hline
\end{tabular}
\end{center}
\caption{Performance analysis of the proposed BrainSegNet over $3$ patients data.}
\label{tab:2}
\end{table}

\subsection{Experimental Results} 
The results with respect to the accuracy values and the recall values according to the testing strategy has been depicted in Table~\ref{tab:2}. As we can see that we have achieved state of the art results in almost all the testing strategies adopted in this paper.  As we see when we move from intra to inter testing strategy the accuracy falls, this is due to the fact that the model has been trained on an entirely other brain and the brain may differ in size or shape and hence may differ slightly in the paths taken from one part to another. One can also observe  that the merged training and testing strategy also gives less accurate results than intra as it is more of a generalized model for accommodating fiber from any one of the brains. For the Table ~\ref{tab:2}, one can infer that the proposed BrainSegNet achieves an accuracy more than $93\%$ and $95\%$ for macro and micro level respectively. Over such a huge and diverse dataset with small training samples we have achieved quit high performance.

\subsection{Comparative Analysis} 

The recall values in Table \ref{tab:2}, signify the superiority of the proposed BrainSegNet in classifying white fibers (which are fewer in number due to sever data imbalance). The proposed BrainSegNet gives far better results than the model proposed in ~\cite{Patel:2016:ABT:3009977.3010013}, in terms of recall values with better accuracy's in most of the experiments. In ~\cite{Patel:2016:ABT:3009977.3010013}, inter brain analysis has not been performed at all, which is very challenging and important to be reported. Even after such a huge data imbalance and intra-class variations the proposed network has achieved state-of-the-art performance and significantly outperforms the existing work ~\cite{Patel:2016:ABT:3009977.3010013}.

\section{Conclusion}
In this paper we propose a novel stacked bidirectional LSTM based segmentation network, (BrainSegNet) for human brain fiber tractography data classification. We perform a two-level hierarchical classification a) White vs Grey matter (Macro) and b) White matter clusters (Micro). Our experimental evaluations show that our model achieves state-of-the-art results. We performed classification at both macro and micro levels that can eliminate the need for manually segmenting the brain fiber tracts which is presently a big issue. We are in the process to get more labeled data, so that we can train another better and deep generalizable network.

\bibliographystyle{plain}
\bibliography{reference}
\end{document}